\documentclass[11pt,a4paper,logo,copyright,nonumbering]{xiaomi}

\usepackage[authoryear,sort&compress,round]{natbib}
\usepackage{makecell}
\usepackage{placeins}
\usepackage{float}
\usepackage{wrapfig}
\hypersetup{colorlinks=true,linkcolor=blue!55!black,citecolor=blue!55!black,urlcolor=blue!55!black}
\newcommand{\cmark}{\ding{51}}
\newcommand{\xmark}{\ding{55}}
\newcommand{\pmark}{\ensuremath{\blacklozenge}}
\definecolor{rwPositive}{RGB}{0,150,90}
\definecolor{rwPartial}{RGB}{255,128,0}
\definecolor{rwNegative}{RGB}{200,50,50}
\newcommand{\relatedworktablecolors}{%
  \renewcommand{\cmark}{\textcolor{rwPositive}{\ding{51}}}%
  \renewcommand{\pmark}{\textcolor{rwPartial}{\ensuremath{\blacklozenge}}}%
  \renewcommand{\xmark}{\textcolor{rwNegative}{\ding{55}}}%
}

\newcommand{\papertitlefirst}{CodeMidas: Scaling Agentic Coding RL}
\newcommand{\papertitlesecond}{Environments from Code Itself}
\newcommand{\papertitle}{\papertitlefirst\space\papertitlesecond}
\title{\papertitle}
\newcommand{\paperauthors}{%
  \parbox[t]{\textwidth}{%
    \centering\normalfont\sffamily\fontsize{11}{15}\selectfont
    Bowen Ye\textsuperscript{1,2}\textsuperscript{*}\quad
    Lei Li\textsuperscript{1,3}\quad
    Shicheng Li\textsuperscript{1}\quad
    Zihao Yue\textsuperscript{1,4}\quad
    Linghao Zhang\textsuperscript{1}\\[3pt]
    Hanglong Lv\textsuperscript{1,2}\quad
    Yuanxin Liu\textsuperscript{1}\quad
    Wenhan Ma\textsuperscript{1,2}\quad
    Hao Tian\textsuperscript{1}\quad
    Rang Li\textsuperscript{1,2}\\[3pt]
    Jinhao Dong\textsuperscript{1,4}\quad
    Yikai Zhao\textsuperscript{1,2}\quad
    Xiangwei Deng\textsuperscript{1,2}\quad
    Hailin Zhang\textsuperscript{1}\\[3pt]
    Liang Zhao\textsuperscript{1}\quad
    Qi Liu\textsuperscript{3}\quad
    Lingpeng Kong\textsuperscript{3}\quad
    Tong Yang\textsuperscript{2,\textdagger}\quad
    Fuli Luo\textsuperscript{1,\textdagger}\\[7pt]
    \textsuperscript{1}LLM Core, Xiaomi\qquad
    \textsuperscript{2}Peking University\\[2pt]
    \textsuperscript{3}University of Hong Kong\qquad
    \textsuperscript{4}Renmin University of China
  }%
}
\hypersetup{pdfauthor={Bowen Ye, Lei Li, Shicheng Li, Zihao Yue, Linghao Zhang, Hanglong Lv, Yuanxin Liu, Wenhan Ma, Hao Tian, Rang Li, Jinhao Dong, Yikai Zhao, Xiangwei Deng, Hailin Zhang, Liang Zhao, Qi Liu, Lingpeng Kong, Tong Yang, Fuli Luo}}

\hypersetup{pdftitle={\papertitle}}

\begin{abstract}
Training capable coding agents via reinforcement learning (RL) requires diverse tasks with reliable verifiers. Open-source codebases offer a rich source of such tasks, while existing methods typically rely on development artifacts such as issues and commits, limiting the range of tasks that can be extracted. To better scale RL environments, we present CodeMidas, an agentic pipeline that turns implemented functionality in existing codebases into executable RL environments using source code as its only task-specific input. CodeMidas allocates agentic compute to every stage of environment construction: agents explore implemented functionality to formulate behavioral specifications, construct tests grounded in execution of the original code, and validate and filter candidate tasks through execution checks and repeated solution rollouts. The resulting dataset has 5,545 training tasks from 3,185 open-source codebases spanning 23 programming languages and 15 technical domains. Training MiMo-V2.5 on these tasks with GRPO improves performance on all five diverse benchmarks, covering
issue repair (DeepSWE + 11.7\%), whole-program construction (ProgramBench +17\%), and terminal work (Terminal-Bench v2.1 +8.5\%). Ablations show that increasing the number of high-quality training tasks improves performance. Trajectory analysis shows the RL-trained agent demonstrates better behaviors like increasing codebase exploration and more diverse self-verification. These results establish source code as a scalable foundation for constructing RL environments that improve coding agents across diverse software tasks.

\end{abstract}

\begin{document}
% Adapted from the custom title block in MiMo_Paper_Template.zip/xiaomi.tex.
\begingroup
\renewcommand{\absfont}{\linespread{1.08}\fontsize{11}{12}\selectfont}
\setlength{\parindent}{0pt}
\vspace*{-8pt}
\begin{adjustwidth}{0pt}{0pt}
\begin{center}
{\titlefont\papertitlefirst\\\papertitlesecond\par}
\vskip7pt
\paperauthors
\end{center}
\end{adjustwidth}
\abscontent
\thispagestyle{firststyle}
\endgroup
\begingroup
\renewcommand{\thefootnote}{\fnsymbol{footnote}}
\footnotetext[1]{Work done during an internship at Xiaomi.}
\footnotetext[2]{Co-corresponding authors.}
\endgroup
\setcounter{footnote}{0}
\par

% Compact only the Introduction paragraph gaps.
\begingroup
\setlength{\parskip}{3pt}
\section{Introduction}\label{sec:intro}

Large language models are increasingly capable of agentic coding: completing substantial pieces of real software work autonomously over long horizons~\citep{sweagent,openhands,swebenchpro}. Building on reinforcement learning (RL) for code generation~\citep{coderl,acecoder}, recent studies have shown substantial gains in real-world software engineering~\citep{swerl,sweuniverse}. Effective RL requires diverse tasks and reliable rewards: task diversity supports generalization~\citep{swesmith}, while reliable rewards help reinforce correct work~\citep{sweuniverse,swerebenchv2}. A central challenge is thus how to turn real codebases into a broad supply of training tasks with trustworthy verifiers.

Existing pipelines construct such environments from development artifacts. Some derive task statements from issues, pull requests, or commits, either directly or via model rewriting~\citep{swebench,r2egym,swerebenchv2,sweuniverse}. Others synthesize faults or development tasks around existing tests~\citep{swesmith,sweflow,swehub}, or use existing documentation to specify the requested functionality~\citep{r2e,mindforge}. These approaches tie task creation to the coverage of recorded changes, tests, or documentation. This motivates building RL environments directly from code: turning implemented functionality into diverse training tasks at scale, each with reliable verifiers.

Open-source codebases provide a rich foundation for this approach, with large code corpora covering hundreds of programming languages~\citep{starcoder2}. Implemented functionality provides both the basis for a task and a candidate solution. Its public interfaces and observable behavior help define what an agent should implement, while executing the original code provides evidence for test expectations. The surrounding codebase can be adapted into a development starting point that preserves real project structure and dependencies. These elements together support the construction of task statements, development environments, and executable verifiers directly from code. Realizing this potential requires making the required behavior explicit in the task statement while leaving internal implementation choices open~\citep{swerebenchv2}. Tests must enforce these requirements, rejecting incorrect solutions while still accepting alternative correct implementations~\citep{testoracle,evalplus,patchdiff,swebenchpro}.

We present \textbf{CodeMidas}, an agentic pipeline that automatically constructs executable coding RL environments using source code as its only task-specific input. CodeMidas identifies existing functionality, formulates behavioral task statements, and adapts codebases into development starting points where the target functionality remains to be implemented. It builds tests informed by execution of the original code and checks execution consistency. CodeMidas then applies post-rollout filtering: adversarial rollouts probe for exploitable leakage, solution reviews assess verifier decisions against the stated requirements, and rollout success rates guide task selection.

Using CodeMidas, we construct 5,545 verifiable training tasks from 3,185 open-source codebases
spanning 23 programming languages and 15 technical domains. We then train MiMo-V2.5 on these tasks using GRPO~\citep{grpo} and observe improvements on all five external benchmarks. Notably, DeepSWE~\citep{deepswe} pass rate rises from 10.0\% to 21.7\%, and the ProgramBench~\citep{programbench} Almost Solved score rises from 4.5 to 21.5. On Terminal-Bench v2.1~\citep{terminalbench}, the pass rate rises from 63.7\% for the initial policy to 72.2\% after RL training. These improvements span issue repair, whole-program construction, code translation, and terminal work, showing that tasks constructed from existing functionality provide strong training signals that transfer across diverse forms of software work.

We further examine how task scale and quality affect these gains, and how agent behavior evolves during training. Increasing the high-quality tasks from 1k to 3k to 5,545 yields progressively higher scores on SWE-bench Pro~\citep{swebenchpro}, DeepSWE, and CodeMidas Val; even the 3k subset outperforms an 8k baseline constructed without cleaning and filtering on all three. As RL progresses, agents explore codebases more and perform more varied self-verification, with agent-written checks associated with higher success rates. These changes also appear on external tasks, providing behavioral evidence of generalization that complements the benchmark gains.

Together, these findings point to source code itself as a basis for scaling coding RL: implemented functionality can be transformed into verifiable learning environments that support generalization across diverse forms of software work. CodeMidas applies a Midas touch to this resource, turning existing code into RL environments improving coding agents across diverse software tasks.

\par
\endgroup
% Keep the opening Related Work headings with text at the foot of page 2.
\begingroup
\titlespacing*{\section}{0pt}{1ex plus 1pt minus 1pt}{3pt}
% CodeMidas manuscript layout applied.
\section{Related Work}\label{sec:related-work}

\subsection{Building coding RL environments}

SWE-bench~\citep{swebench} established repository-level issue resolution as an execution-based evaluation setting. Later pipelines scale task collection from issues and pull requests~\citep{swerebenchv2,sweuniverse,davincienv,scaleswe,swenext}. R2E-Gym~\citep{r2egym} generates tests and task statements from commits, reducing reliance on human-written issues and tests. These methods seed tasks from development records.

Other pipelines build tasks around existing tests or documentation. SWE-smith~\citep{swesmith} synthesizes code changes that break existing tests, while SWE-Flow~\citep{sweflow} derives incremental development tasks from unit tests and their runtime dependencies. SWE-Hub~\citep{swehub} combines test-validated bug synthesis with repository construction based on coverage and structured requirements. R2E~\citep{r2e} refines function docstrings into specifications, and MindForge~\citep{mindforge} exposes documentation and compiled reference programs for from-scratch implementation. Table~\ref{tab:related-work} lists their task-specific input requirements. CodeMidas uses source code as its only task-specific input, deriving behavioral statements and execution-grounded tests beyond the coverage of development records, documentation, and existing tests.

\begin{table}[!htbp]
\relatedworktablecolors
\centering
\caption{Task-specific input requirements and language coverage of representative pipelines for constructing coding environments. \cmark{} = not required; \pmark{} = required by part of the pipeline; \xmark{} = required.}
\label{tab:related-work}
\footnotesize
\setlength{\tabcolsep}{3pt}
\renewcommand{\arraystretch}{1.15}
\begin{tabularx}{\linewidth}{@{}>{\raggedright\arraybackslash}X*{5}{c}r@{}}
\toprule
\textbf{Method} & \textbf{\makecell{w/o\\issue}} & \textbf{\makecell{w/o\\PR}} & \textbf{\makecell{w/o\\commit}} & \textbf{\makecell{w/o existing\\tests}} & \textbf{\makecell{w/o written\\description}} & \textbf{Languages} \\
\midrule
SWE-rebench V2~\citep{swerebenchv2} & \pmark{} & \xmark{} & \xmark{} & \xmark{} & \cmark{} & 20 \\
daVinci-Env~\citep{davincienv} & \xmark{} & \xmark{} & \xmark{} & \xmark{} & \cmark{} & 1 \\
R2E-Gym~\citep{r2egym} & \cmark{} & \cmark{} & \xmark{} & \pmark{} & \cmark{} & 1 \\
SWE-smith~\citep{swesmith} & \cmark{} & \pmark{} & \pmark{} & \xmark{} & \cmark{} & 1 \\
SWE-Flow~\citep{sweflow} & \cmark{} & \cmark{} & \cmark{} & \xmark{} & \cmark{} & 1 \\
SWE-Hub~\citep{swehub} & \cmark{} & \cmark{} & \cmark{} & \xmark{} & \pmark{} & 11 \\
R2E~\citep{r2e} & \cmark{} & \cmark{} & \cmark{} & \cmark{} & \xmark{} & 1 \\
MindForge~\citep{mindforge} & \cmark{} & \cmark{} & \cmark{} & \cmark{} & \xmark{} & 15 \\
\textbf{CodeMidas} & \textbf{\cmark{}} & \textbf{\cmark{}} & \textbf{\cmark{}} & \textbf{\cmark{}} & \textbf{\cmark{}} & \textbf{23} \\
\bottomrule
\end{tabularx}
\end{table}

\Needspace{11\baselineskip}
\subsection{Rewards and verification for coding agents}

CodeRL~\citep{coderl} combines unit-test feedback with a learned critic. At repository level, SWE-RL~\citep{swerl} uses reference-patch similarity, while SWE-Universe~\citep{sweuniverse} trains agents in executable environments. AceCoder~\citep{acecoder} synthesizes tests to study learned rewards and direct test-pass rewards. CodeMidas uses GRPO~\citep{grpo} with execution rewards from synthesized tests, without a reward model or learned verifier.

CodeT~\citep{codet} selects programs using generated tests and execution agreement. For SWE agents, SWE-Shepherd~\citep{sweshepherd} scores actions with a process reward model, Agentic Rubrics~\citep{agenticrubrics} scores patches against codebase-grounded rubrics without test execution, and R2E-Gym~\citep{r2egym} combines learned and execution-based verifiers. Self-Debugging~\citep{selfdebugging} and Reflexion~\citep{reflexion} use execution feedback or verbal reflection to revise solutions across attempts. Our trajectory analysis examines agents' exploration and self-verification during RL training and on held-out tasks.

Reliable execution rewards also depend on the test oracle~\citep{testoracle}. EvalPlus~\citep{evalplus} shows that expanded tests uncover incorrect generated programs missed by original suites, and PatchDiff~\citep{patchdiff} documents incorrect patches accepted by SWE-bench tests. SWE-bench Pro~\citep{swebenchpro} and SWE-rebench V2~\citep{swerebenchv2} discuss specification gaps and overly restrictive tests. CodeMidas combines execution consistency checks with post-rollout filtering: adversarial rollouts probe leakage, solution reviews check test verdicts, and rollout outcome filtering retains tasks with both successful and failed attempts.

\par
\endgroup
% CodeMidas manuscript layout applied.
\begin{figure}[!t]
\centering
\includegraphics[width=\linewidth]{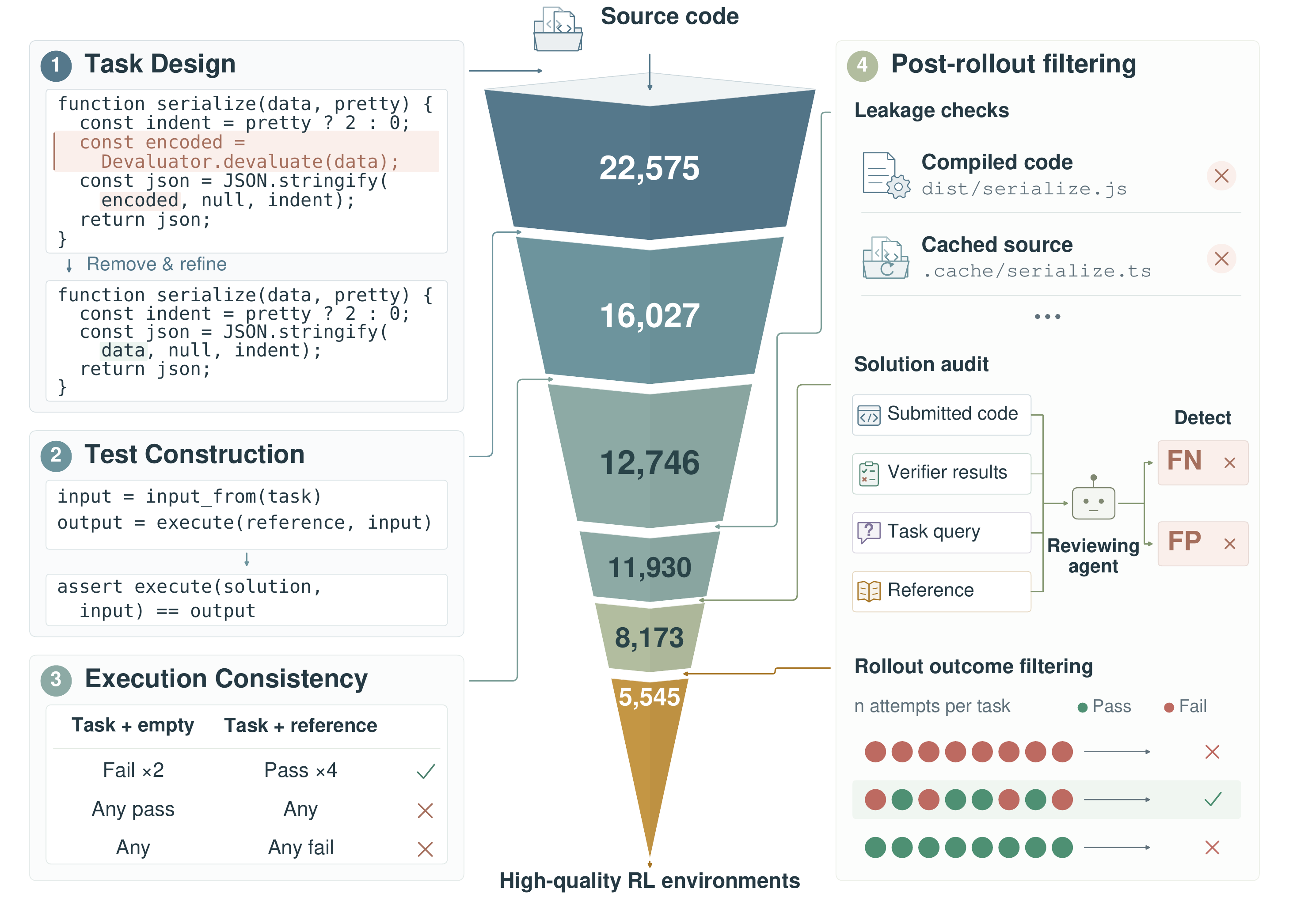}
\caption{\textbf{CodeMidas overview.} Four modules cover task design, test construction, execution consistency, and post-rollout filtering. The pyramid shows retained task counts respectly.}
\label{fig:method-overview}
\end{figure}

\section{Method}\label{sec:method}

In this section, we describe how CodeMidas constructs and filters coding RL environments using source code as its only task-specific input (Figure~\ref{fig:method-overview}). We first introduce task design and codebase adaptation (\S\ref{sec:task-construction}), followed by execution-grounded test construction (\S\ref{sec:verifier-construction}), and environment preparation with execution consistency check (\S\ref{sec:environment}). We then describe how agent rollouts are used to filter environments before RL training (\S\ref{sec:validation}), and summarize the resulting dataset (\S\ref{sec:dataset-overview}).

Each task consists of a statement, a containerized development environment, and a hidden executable verifier. The solver receives the statement and adapted codebase with its dependencies. Throughout solving, the verifier is kept outside the solver's environment. It is injected only at grading to evaluate the completed implementation and return a binary execution reward for RL.

\subsection{Task Design and Codebase Adaptation}\label{sec:task-construction}

An agent inspects codebase structure and build metadata to identify functionality with public entry points and observable outcomes. We prioritize tasks requiring reasoning across the codebase. Supported interfaces include command-line tools, pure library functions, and stateful library APIs, assessed through process outputs, return values, and state changes across calls.

For each candidate, the agent traces public entry points and shared dependencies to define the task scope. It removes the selected core implementation, then adjusts the remaining code to form a coherent starting point for the requested work. The task statement and code boundaries are revised together while preserving shared components and project context. The original implementation is retained separately to provide a reference solution for the task.

The statement defines inputs, observable behavior, and required public interfaces. Solvers implement the missing codebase functionality, choosing their own internal helpers and algorithms.

\subsection{Execution-grounded Test Construction}\label{sec:verifier-construction}

An agent maps the task statement's behavioral requirements to test inputs and boundary cases, invokes public entry points in a reference copy of the codebase, and records the outcomes. Tests use command executions for CLI tools, input-output cases for pure functions, and sequences of calls for stateful APIs. Stateful tests exercise dependencies across calls, including ordering and cleanup behavior when specified. Each test records the specific requirement that it covers.

For outputs and properties fixed by the statement, assertions use reference execution to establish expected values. For aspects left unspecified, assertions check only the stated constraints. For example, tests enforce a required exception type without fixing unspecified message wording. Cases with distinct expected outputs probe input-dependent behavior.

An agent then reviews every assertion for restrictions unsupported by the statement, such as exact wording, incidental ordering, or internal structure. It replaces these restrictions with behavioral checks while preserving the checks required by the statement. A task is rejected if an assertion depends on a private symbol and has no behavioral substitute. The revised tests are rerun on the reference solution to confirm that they remain compatible. After review, test inputs and assertions are fixed for grading, which runs the submitted implementation against these checks.

\subsection{Environment Preparation}\label{sec:environment}

\noindent \textbf{Environment preparation.} Starting from a uniform base container image, an agent installs dependencies and prepares build and runtime resources according to the project's declarations. Cleanup removes artifacts that could reveal the deleted implementation, including compiled outputs, cached copies, and files left by construction agents. Original tests related to the target functionality are also removed. Required packages, fixtures, and build wrappers are retained to support building and running completed implementations in the prepared environment.

\noindent \textbf{Execution consistency.} Each task is checked under the training runtime settings in six fresh containers: two with the starting codebase and four with the reference solution in place. Both starting-state runs must fail and all four reference runs must pass. These repetitions check the expected fail-to-pass transition and screen for unstable execution outcomes.

\begin{figure}[!t]
\centering
\begin{minipage}[t]{0.485\linewidth}
\vspace{0pt}
\centering
\includegraphics[width=\linewidth]{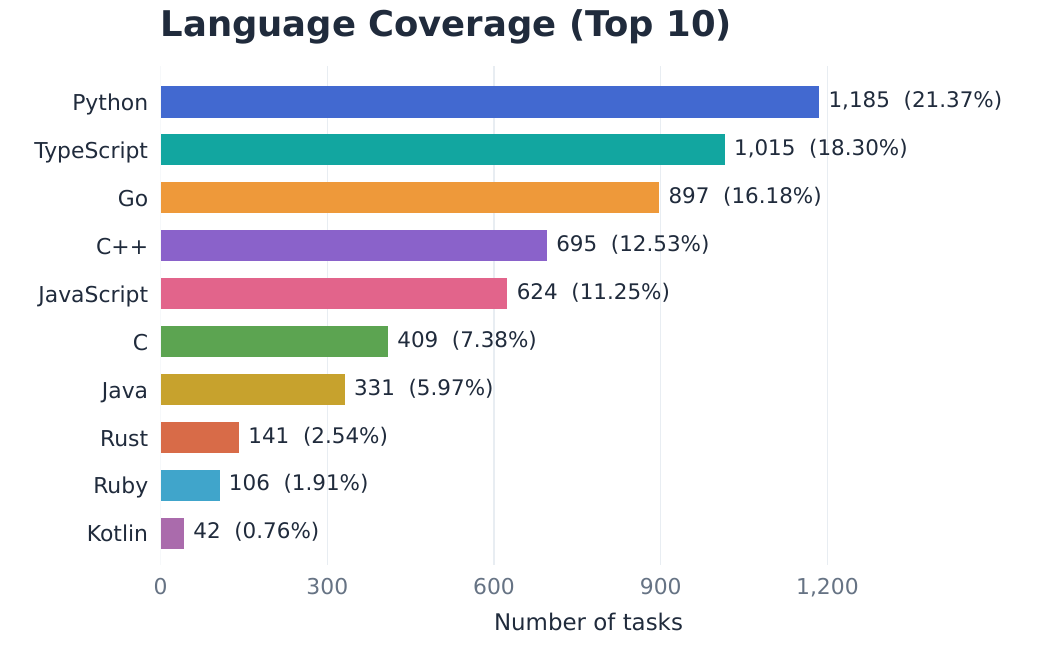}
% Match the adjacent domain plot's height so both captions start on the same line.
\par\vspace{10.62pt}
\caption{\textbf{Language coverage of the training set.} Tasks inherit the primary language label of their codebase. The ten most frequent languages cover 5,445 of 5,545 tasks (98.2\%); the full dataset spans 23 languages. Labels report task counts and the associated percentages of the full training set.}
\label{fig:dataset-languages}
\end{minipage}
\hfill
\begin{minipage}[t]{0.485\linewidth}
\vspace{0pt}
\centering
\includegraphics[width=\linewidth]{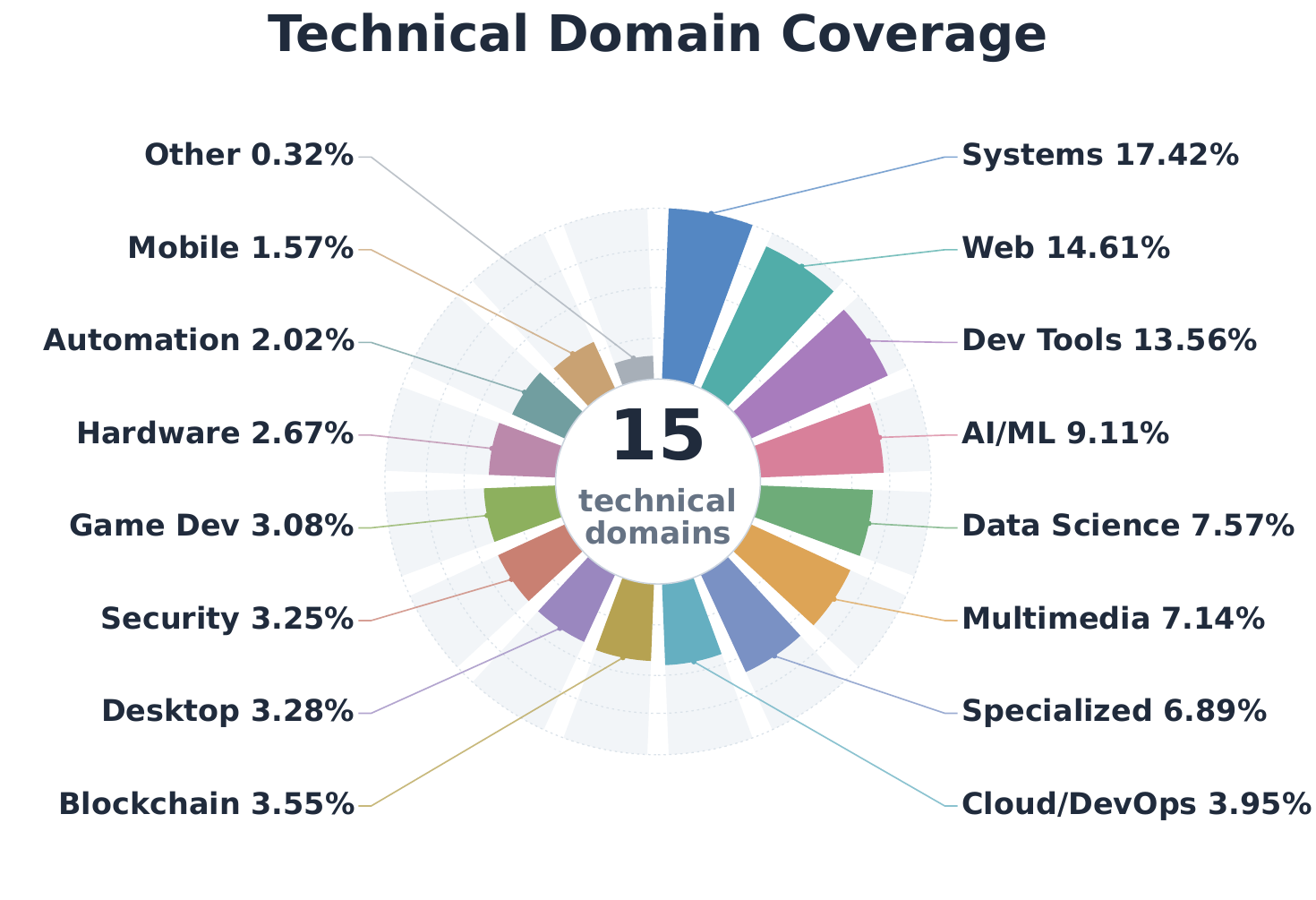}
\caption{\textbf{Technical domain coverage of the training set.} The 15 labeled domains and Other (18 unlabeled tasks, 0.32\%) are ordered clockwise by decreasing task share. Domain labels come from codebases. Radial bar length is proportional to the square root of task share respectively.}
\label{fig:dataset-domains}
\end{minipage}
\end{figure}

\subsection{Post-rollout Environment Filtering}\label{sec:validation}

Execution checks cover the starting codebase and the reference solution. Before RL training, we further filter environments using agent rollouts and their outcomes. Post-rollout filtering checks for exploitable leakage, disagreement between solution assessments and test verdicts, and tasks for which all attempts pass or all attempts fail under the screening model.

\noindent \textbf{Leakage filtering.} In adversarial rollouts, an agent tries to exploit residual leakage to recover a solution without doing the intended development work. It searches the full solver-visible environment, including compiled artifacts, caches, files left by construction agents, and installed copies of the target project. It logs the commands and outputs supporting each suspected exploit. A separate review checks the evidence against the reference solution and verifier. We reject tasks if the review confirms that leaked material can bypass the intended implementation work.

\noindent \textbf{Agreement on agent solutions.} To assess the verifier on agent-generated implementations, a coding agent tries four times per task. A reviewing agent examines the resulting rollout trajectories, including submitted
code and test outputs, alongside the task statement, verifier, and reference solution. Using this evidence, it assesses whether each implementation satisfies the statement and checks for mismatches between the stated requirements and the verifier’s behavior. It flags false positives when an implementation judged incorrect passes the tests, and false negatives when an implementation judged correct fails. Tasks with identified verifier defects are rejected.

\noindent \textbf{Rollout outcome filtering.} In another check, a frontier model makes several attempts per task, scored by the verifier. All-pass or all-fail outcomes may reflect task difficulty or remaining defects, such as weak tests or requirements missing from the statement. These outcomes do not reveal the cause. We keep only tasks with both successful and failed attempts under this model and budget.

\begingroup
\setlength{\textfloatsep}{8pt}
\subsection{Dataset Overview}\label{sec:dataset-overview}

We retain 5,545 tasks from 3,185 codebases across 23 languages and 15 technical domains.

\begin{figure}[!b]
\centering
\includegraphics[width=0.52\linewidth]{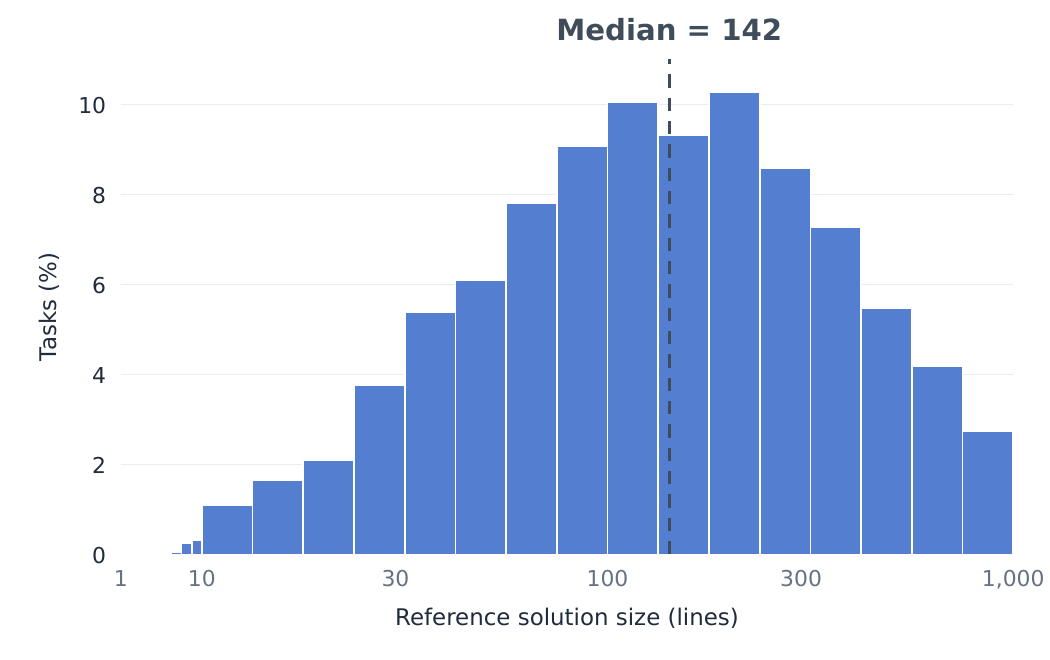}
\caption{\textbf{Reference solution size.} Bars show task percentages; the dashed line marks the median (142 lines). The piecewise log x-axis compresses the 1--10 interval to one fifth the width of subsequent decades.}
\label{fig:dataset-reference-size}
\end{figure}

\noindent \textbf{Language and domain coverage.} Figures~\ref{fig:dataset-languages} and~\ref{fig:dataset-domains} summarize the dataset’s coverage across 23
programming languages and 15 technical domains. Python (21.4\%), TypeScript (18.3\%), and Go (16.2\%) are the most represented
languages, followed by C++ (12.5\%) and JavaScript (11.3\%). Systems
software (17.4\%), web technologies (14.6\%), and developer tools (13.6\%) are the largest technical
domains, together accounting for 45.6\% of tasks. 

\noindent \textbf{Reference solution size.} We count all source lines added or deleted in the reference patch, including comments and blank lines. Across the full training set, the median is 142 lines, with an interquartile range of 66--305 lines. Reference patches touch at least two source files in 65.9\% of tasks. Figure~\ref{fig:dataset-reference-size} groups reference solution sizes into equal log-width bins and reports the percentage of all 5,545 training tasks represented in each of these bins.

\FloatBarrier
\endgroup

% CodeMidas manuscript layout applied.
\Needspace{5\baselineskip}
\section{Experiments}\label{sec:experiments}

In this section, we describe our training and evaluation setup (Section~\ref{sec:experimental-setup}). We then present results on external benchmarks and examine learning dynamics on CodeMidas Val (Section~\ref{sec:results}).

% Keep the benchmark overview immediately before the setup section.
\begingroup
\setlength{\intextsep}{4pt}
\begin{figure}[H]
\centering
\includegraphics[width=0.94\linewidth]{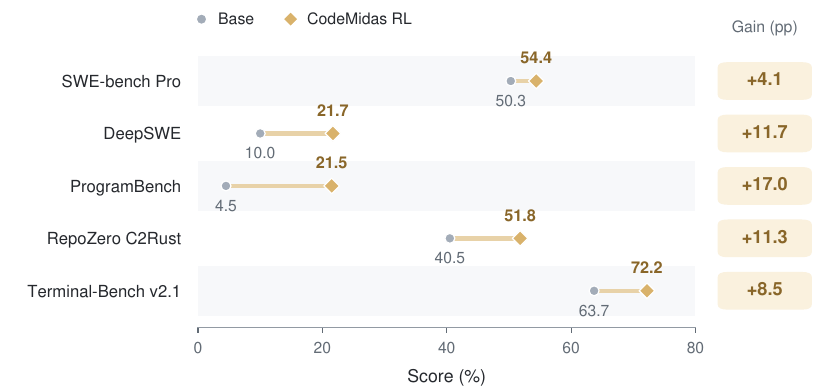}
\caption{\textbf{Performance gains from RL on CodeMidas.} Initial MiMo-V2.5 scores (gray) and CodeMidas RL scores (gold). ProgramBench reports Almost Solved; all other evaluations report pass rate. Labels on the right give absolute improvements over the initial policy in percentage points.}
\label{fig:main-results}
\end{figure}
\endgroup

\subsection{Experimental Setup}\label{sec:experimental-setup}

We train MiMo-V2.5~\citep{mimov25} on 5,545 CodeMidas tasks with GRPO~\citep{grpo}, binary execution rewards, batch size 32, and 32 rollouts per task. See Appendix~\ref{app:training-configuration}.

We use identical evaluation settings for the initial policy and RL checkpoints. We follow the official task sets for the five external benchmarks: SWE-bench Pro~\citep{swebenchpro}, DeepSWE v1.1~\citep{deepswe}, ProgramBench~\citep{programbench}, RepoZero C2Rust~\citep{repozero}, and Terminal-Bench v2.1~\citep{terminalbench}. CodeMidas Val consists of 200 randomly sampled CodeMidas tasks separate from the 5,545 training tasks, with three evaluation attempts per task. We verified that the training set is disjoint from CodeMidas Val and all five external benchmark task sets. For ProgramBench, we report Almost Solved, the percentage of tasks passing at least 95\% of their tests; all other evaluations report pass rate. For each benchmark, we report absolute score improvements relative to the initial policy in percentage points.

\FloatBarrier
\Needspace{20\baselineskip}
\subsection{Results}\label{sec:results}

% Start a paragraph before wrapfig; titlesec otherwise consumes the figure box.
\vspace{-\baselineskip}
\noindent
% Resume full-width text immediately below the five-line caption.
\begin{wrapfigure}[15]{r}{0.50\textwidth}
\centering
\includegraphics[width=\linewidth]{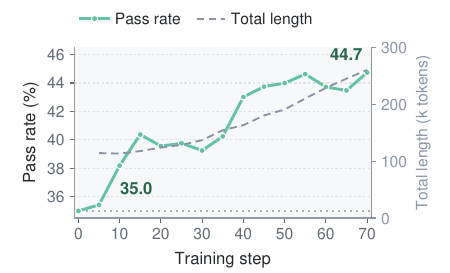}
\caption{\looseness=-1 \textls[-10]{\textbf{Learning curve on CodeMidas Val.} Pass rate (green, left axis) and mean total length (gray-blue dashed, right axis) during RL on CodeMidas. Length is measured in thousands of tokens; the horizontal dotted line marks the pass rate achieved by the initial policy.}}
\label{fig:learning-dynamics}
\end{wrapfigure}

\noindent \textbf{RL improves performance across task types.} Training on CodeMidas improves performance on all five external benchmarks (Figure~\ref{fig:main-results}). DeepSWE pass rate increases from 10.0\% to 21.7\%, while Terminal-Bench v2.1 improves from 63.7\% to 72.2\%. On ProgramBench, the Almost Solved score rises from 4.5 to 21.5. The gains span repository repair, code translation, program construction, and terminal work, supporting the use of source-derived functionality tasks to train \mbox{agents for diverse software work.}

\noindent \textbf{Learning dynamics on CodeMidas Val.} Figure~\ref{fig:learning-dynamics} tracks pass rate and mean total token length during RL on CodeMidas. Pass rate rises from 35.0\% to 44.7\%, staying roughly 8--10 percentage points above the initial rate at evaluated checkpoints from step 40 onward. These gains accompany longer trajectories, indicating greater use of the available interaction budget.
\par\WFclear

% CodeMidas manuscript layout applied.
\section{Analysis}\label{sec:analysis}

% A small tracking adjustment avoids a short fourth line without rewording.
\looseness=-1
\textls[-5]{To understand the gains from training on CodeMidas, we examine the contributions of task scale and quality (Section~\ref{sec:task-scale-quality}). We then analyze how agent behavior changes during RL, how these behaviors are associated with task success, and whether they generalize across task types (Section~\ref{sec:behavior}).}

\subsection{Task Scale and Quality}\label{sec:task-scale-quality}

% Pair the learning curves with the task-pool setup on the same page.
\begin{wrapfigure}[18]{r}{0.50\textwidth}
\centering
\includegraphics[width=\linewidth]{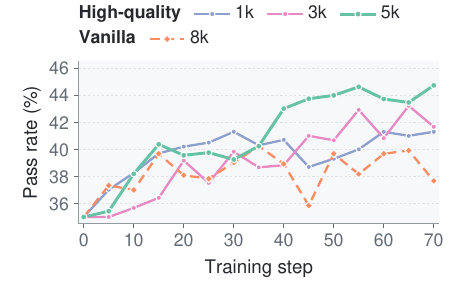}
\caption{\textbf{Learning curves on CodeMidas Val.} Pass rates at five-step intervals from step 0 to 70, without smoothing. Solid lines with circles denote the high-quality 1k, 3k, and full (5k) pools; the dashed line with diamonds denotes the vanilla 8k sample \mbox{drawn before filtering and cleaning.}}
\label{fig:test-learning-curves}
\end{wrapfigure}

To examine the effects of task scale and quality, we train on the full high-quality CodeMidas dataset of 5,545 tasks (5k) and random 1k and 3k subsets. We also train on approximately 8,000 tasks (8k) sampled before filtering, each with a task statement, development environment, and verifier. This vanilla 8k sample is used without environment cleaning or execution consistency checks (Section~\ref{sec:environment}), or any of the three post-rollout filtering \mbox{steps described in Section~\ref{sec:validation}.}

All four settings use identical training configurations and are evaluated over the same checkpoint range. We first examine learning curves on CodeMidas Val, then compare scores on SWE-bench Pro, DeepSWE, and CodeMidas Val for each of the four training settings.

\par\WFclear

\noindent \textbf{The full dataset leads across later checkpoints.} We first examine every evaluated checkpoint on CodeMidas Val (Figure~\ref{fig:test-learning-curves}). The 1k setting reaches 41.30 at step 30, while the 3k setting reaches 43.22 at step 65. The full dataset leads at every evaluated checkpoint from step 40 through step 70, where it reaches 44.73. Its advantage thus persists across the later part of training.

\begin{figure}[!htbp]
\centering
\includegraphics[width=0.90\linewidth]{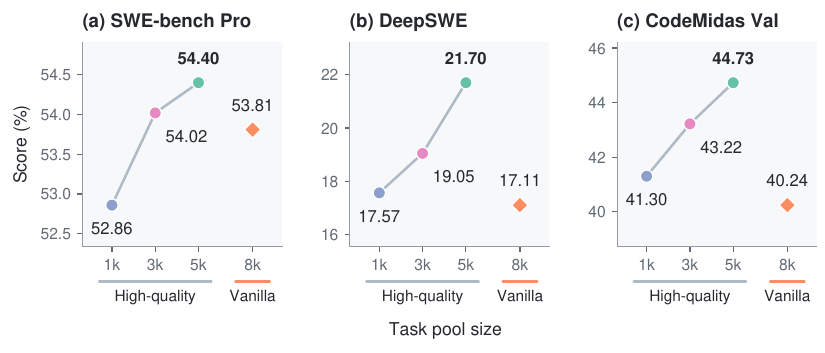}
\caption{\looseness=-1 \textbf{Task scale and quality.} Scores on SWE-bench Pro, DeepSWE, and CodeMidas Val. Connected circles denote high-quality 1k, 3k, and 5k pools; diamonds denote vanilla 8k. Panels use different y-axis ranges.}
\label{fig:quality-scaling}
\end{figure}

\noindent \textbf{Performance improves with task scale.} We next compare scores across all three evaluations (Figure~\ref{fig:quality-scaling}). The 1k, 3k, and 5k pools achieve progressively higher scores: DeepSWE scores rise from 17.57 to 19.05 to 21.70, while scores on CodeMidas Val increase from 41.30 to 43.22 to 44.73. These results support scaling high-quality training data.

\noindent \textbf{The high-quality 5k pool outperforms the vanilla 8k sample.} The full CodeMidas dataset exceeds the vanilla 8k sample by 0.59, 4.59, and 4.49 percentage points on SWE-bench Pro, DeepSWE, and CodeMidas Val, respectively. Even the high-quality 3k subset outperforms the vanilla 8k sample on all three evaluations. These comparisons support the combined value of environment reliability and training suitability provided by cleaning, execution checks, and post-rollout filtering, even with fewer tasks and the same training configuration.
\par\WFclear

\FloatBarrier
\subsection{Behavioral Changes and Generalization}\label{sec:behavior}

\begin{figure}[!htbp]
\centering
\includegraphics[width=\linewidth]{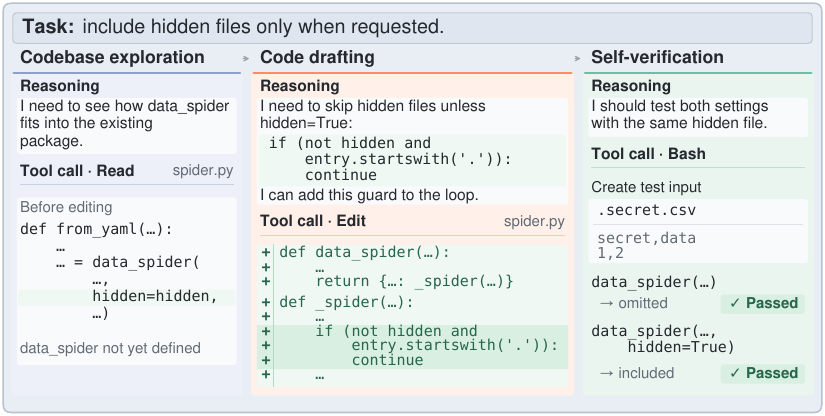}
\caption{\looseness=-1 \textls[-10]{\textbf{Codebase exploration, code drafting, and self-verification in one rollout.} Columns pair reasoning with tool use from one CodeMidas training rollout: reading the caller,
drafting hidden-file filtering and applying it via \texttt{Edit}, and testing both flag settings
with \texttt{.secret.csv}. The diff marks additions; \texttt{Passed} indicates agreement with
expected behavior.}}
\label{fig:behavior-example}
\end{figure}

To characterize the behavioral changes accompanying the performance gains, we analyze codebase exploration, reasoning before code edits, and self-verification (Figure~\ref{fig:behavior-example}). For reasoning before edits, we measure the fraction of written code fragments that appeared in preceding reasoning.

\looseness=-1
\noindent \textls[-10]{\textbf{Exploration and drafting increase; self-verification diversifies.} To quantify behavioral changes during RL, we compare early and late training rollouts (Table~\ref{tab:training-behaviors}). We measure exploration by read/search calls before the first edit, drafting by the fraction of code fragments in \texttt{Write}/\texttt{Edit} payloads already present in preceding reasoning, and self-verification by distinct verification commands after the final repository edit. Exploration increases from 27.2 to 40.1 calls, the drafting ratio from 0.36 to 0.63, and verification commands from 2.03 to 2.53. Agents thus explore more, show greater overlap between written code and preceding reasoning, and execute more diverse verification commands. Definitions of exploration, code drafting, and self-verification are given in Appendix~\ref{app:trajectory-analysis}.}

\noindent \textbf{Self-verification is associated with higher pass rates.} Within the same task and checkpoint on CodeMidas Val, rollouts with checks written and executed by the agent have a mean pass rate 4.2 percentage points higher than those without (95\% CI: 1.8--6.6). Splitting exploration and drafting at checkpoint medians, excluding undefined measurements, gives differences of +0.7 points (95\% CI: -1.9 to 3.7) and +1.95 points (-0.04 to 3.96), respectively. Confidence intervals resample whole tasks, keeping all observed checkpoints from each sampled task together.

\begin{table}[H]
\centering
\caption{\textbf{Changes during RL.} Rollout means in early and late training, using defined measurements for each behavior. Changes are computed from unrounded means for each behavioral measure.}
\label{tab:training-behaviors}
\small
\setlength{\tabcolsep}{3pt}
\renewcommand{\arraystretch}{1.15}
\begin{tabular*}{\linewidth}{@{\extracolsep{\fill}}llrrr@{}}
\toprule
\textbf{Behavior} & \textbf{Measure} & \textbf{Early} & \textbf{Late} & \textbf{Change} \\
\midrule
Codebase exploration & Pre-edit read/search calls & 27.2 & 40.1 & +12.9 \\
Code drafting & Drafting ratio & 0.358 & 0.629 & +0.271 \\
Self-verification & Distinct post-edit commands & 2.03 & 2.53 & +0.50 \\
\bottomrule
\end{tabular*}
\end{table}

\looseness=-1
\noindent \textbf{Behavioral changes generalize; interaction length varies by task.} To assess generalization beyond CodeMidas, we compare early and late checkpoints on SWE-bench Pro, ProgramBench, and Terminal-Bench v2.1 (Table~\ref{tab:held-out-patterns}). We also measure interaction length in assistant turns to examine how it changes across task types. Exploration increases across all three benchmarks, while changes in verification diversity vary by benchmark. Drafting ratios increase on SWE-bench Pro and ProgramBench. Mean interaction length increases from 37.3 to 50.1 turns for issue repair on SWE-bench Pro and decreases from 155.1 to 122.8 for whole-program construction on ProgramBench. On ProgramBench, greater codebase exploration thus accompanies shorter overall interactions.

\begin{table}[H]
\centering
\caption{\textbf{Behavior on held-out tasks.} Means over the first \(\rightarrow\) last three observed checkpoints for each benchmark. The three behavioral measures follow Table~\ref{tab:training-behaviors} and include rollouts with defined measurements; interaction length includes all scored rollouts. Measured subsets can vary across checkpoints.}
\label{tab:held-out-patterns}
\small
\setlength{\tabcolsep}{3pt}
\renewcommand{\arraystretch}{1.15}
\begin{tabular*}{\linewidth}{@{\extracolsep{\fill}}lrrrr@{}}
\toprule
\textbf{Evaluation} & \makecell[r]{\textbf{Codebase exploration}\\(read/search calls)} & \makecell[r]{\textbf{Code drafting}\\(drafting ratio)} & \makecell[r]{\textbf{Self-verification}\\(distinct commands)} & \makecell[r]{\textbf{Interaction length}\\(assistant turns)} \\
\midrule
SWE-bench Pro & 23.1 \(\rightarrow\) 35.5 & 0.304 \(\rightarrow\) 0.653 & 0.80 \(\rightarrow\) 0.96 & 37.3 \(\rightarrow\) 50.1 \\
ProgramBench & 55.7 \(\rightarrow\) 83.6 & 0.106 \(\rightarrow\) 0.361 & 0.93 \(\rightarrow\) 0.99 & 155.1 \(\rightarrow\) 122.8 \\
Terminal-Bench v2.1 & 11.9 \(\rightarrow\) 16.8 & N/A & 2.01 \(\rightarrow\) 2.61 & 59.2 \(\rightarrow\) 69.5 \\
\bottomrule
\end{tabular*}
\end{table}

\Needspace{13\baselineskip}
\section{Conclusion}\label{sec:conclusion}

CodeMidas demonstrates the value of existing codebases as a source of coding RL data. Training on 5,545 tasks improves MiMo-V2.5 across five external benchmarks, with gains spanning diverse forms of software work. Analyses show that expanding high-quality task pools improves performance and that a smaller filtered dataset can outperform a larger unfiltered one. During training, agents explore codebases more and perform more varied self-verification; agent-written checks are associated with higher success rates, and these changes also appear on external tasks.

\looseness=-1
CodeMidas applies a Midas touch to existing codebases, turning implemented functionality into effective training environments and providing a practical route to scaling coding RL from code itself.

\FloatBarrier
\begingroup
\setlength{\bibsep}{2pt plus 1pt minus 1pt}
\bibliographystyle{abbrvnat}
\bibliography{references}

@inproceedings{acecoder,
  title = {{ACECODER: Acing Coder RL via Automated Test-Case Synthesis}},
  author = {Huaye Zeng and Dongfu Jiang and Haozhe Wang and Ping Nie and Xiaotong Chen and Wenhu Chen},
  booktitle = {Proceedings of the 63rd Annual Meeting of the Association for Computational Linguistics (Volume 1: Long Papers)},
  pages = {12023--12040},
  publisher = {Association for Computational Linguistics},
  year = {2025},
  doi = {10.18653/v1/2025.acl-long.587},
  url = {https://aclanthology.org/2025.acl-long.587/},
}

@inproceedings{agenticrubrics,
  title = {{Agentic Rubrics as Contextual Verifiers for SWE Agents}},
  author = {Mohit Raghavendra and Anisha Gunjal and Bing Liu and Yunzhong He},
  booktitle = {Proceedings of the 64th Annual Meeting of the Association for Computational Linguistics (Volume 1: Long Papers)},
  pages = {15265--15290},
  publisher = {Association for Computational Linguistics},
  year = {2026},
  doi = {10.18653/v1/2026.acl-long.697},
  url = {https://aclanthology.org/2026.acl-long.697/},
}

@misc{davincienv,
  title = {{daVinci-Env: Open SWE Environment Synthesis at Scale}},
  author = {Dayuan Fu and Shenyu Wu and Yunze Wu and Zerui Peng and Yaxing Huang and Jie Sun and Ji Zeng and Mohan Jiang and Lin Zhang and Yukun Li and Jiarui Hu and Liming Liu and Jinlong Hou and Pengfei Liu},
  year = {2026},
  url = {https://arxiv.org/abs/2603.13023},
  eprint = {2603.13023},
  archivePrefix = {arXiv},
  primaryClass = {cs.SE},
  howpublished = {arXiv preprint arXiv:2603.13023},
}

@misc{deepswe,
  title = {{DeepSWE v1.1}},
  author = {Wenqi Huang and Peter Jiang},
  year = {2026},
  url = {https://deepswe.datacurve.ai/blog/deepswe-v1-1},
}

@misc{grpo,
  title = {{DeepSeekMath: Pushing the Limits of Mathematical Reasoning in Open Language Models}},
  author = {Zhihong Shao and Peiyi Wang and Qihao Zhu and Runxin Xu and Junxiao Song and Xiao Bi and Haowei Zhang and Mingchuan Zhang and Y. K. Li and Y. Wu and Daya Guo},
  year = {2024},
  url = {https://arxiv.org/abs/2402.03300},
  eprint = {2402.03300},
  archivePrefix = {arXiv},
  primaryClass = {cs.CL},
  howpublished = {arXiv preprint arXiv:2402.03300},
}

@misc{mindforge,
  title = {{MindForge: Teaching Small Language Models Whole-Life-Cycle Software Engineering via Source-Free Program Synthesis}},
  author = {Yihao Chen and Shi Chang and Khaled Chawa and Feng Lin and Boyuan Chen and Shaowei Wang and Ahmed E. Hassan},
  year = {2026},
  url = {https://arxiv.org/abs/2607.27146},
  eprint = {2607.27146},
  archivePrefix = {arXiv},
  primaryClass = {cs.SE},
  howpublished = {arXiv preprint arXiv:2607.27146},
}

@misc{mimov25,
  title = {{MiMo-V2.5}},
  author = {{Xiaomi MiMo Team}},
  year = {2026},
  url = {https://mimo.xiaomi.com/mimo-v2-5},
}

@misc{programbench,
  title = {{ProgramBench: Can Language Models Rebuild Programs From Scratch?}},
  author = {John Yang and Kilian Lieret and Jeffrey Ma and Parth Thakkar and Dmitrii Pedchenko and Sten Sootla and Emily McMilin and Pengcheng Yin and Rui Hou and Gabriel Synnaeve and Diyi Yang and Ofir Press},
  year = {2026},
  url = {https://arxiv.org/abs/2605.03546},
  eprint = {2605.03546},
  archivePrefix = {arXiv},
  primaryClass = {cs.SE},
  howpublished = {arXiv preprint arXiv:2605.03546},
}

@inproceedings{r2e,
  title = {{R2E: Turning any Github Repository into a Programming Agent Environment}},
  author = {Naman Jain and Manish Shetty and Tianjun Zhang and King Han and Koushik Sen and Ion Stoica},
  booktitle = {Proceedings of the 41st International Conference on Machine Learning},
  volume = {235},
  series = {Proceedings of Machine Learning Research},
  pages = {21196--21224},
  publisher = {PMLR},
  year = {2024},
  url = {https://proceedings.mlr.press/v235/jain24c.html},
}

@inproceedings{r2egym,
  title = {{R2E-Gym: Procedural Environment Generation and Hybrid Verifiers for Scaling Open-Weights SWE Agents}},
  author = {Naman Jain and Jaskirat Singh and Manish Shetty and Tianjun Zhang and Liang Zheng and Koushik Sen and Ion Stoica},
  booktitle = {Second Conference on Language Modeling},
  year = {2025},
  url = {https://openreview.net/forum?id=7evvwwdo3z},
}

@misc{repozero,
  title = {{RepoZero: Can LLMs Generate a Code Repository from Scratch?}},
  author = {Zhaoxi Zhang and Yiming Xu and Jiahui Liang and Weikang Li and Xiaoshuai Chen and Liwei Qian and Xin Pei and Jizhou Huang and Run Sun and Yunfang Wu},
  year = {2026},
  url = {https://arxiv.org/abs/2605.07122},
  eprint = {2605.07122},
  archivePrefix = {arXiv},
  primaryClass = {cs.SE},
  howpublished = {arXiv preprint arXiv:2605.07122},
}

@misc{scaleswe,
  title = {{Immersion in the GitHub Universe: Scaling Coding Agents to Mastery}},
  author = {Jiale Zhao and Guoxin Chen and Fanzhe Meng and Minghao Li and Jie Chen and Hui Xu and Yongshuai Sun and Wayne Xin Zhao and Ruihua Song and Yuan Zhang and Peng Wang and Cheng Chen and Jirong Wen and Kai Jia},
  year = {2026},
  url = {https://arxiv.org/abs/2602.09892},
  eprint = {2602.09892},
  archivePrefix = {arXiv},
  primaryClass = {cs.SE},
  howpublished = {arXiv preprint arXiv:2602.09892},
}

@inproceedings{swebenchpro,
  title = {{SWE-Bench Pro: Can AI Agents Solve Long-Horizon Software Engineering Tasks?}},
  author = {Xiang Deng and Jeff Da and Edwin Pan and Yannis Yiming He and Charles Ide and Kanak Garg and Niklas Lauffer and Andrew Park and Chetan Rane and Karmini Sampath and Maya Krishnan and Srivatsa Kundurthy and Sean Hendryx and Zifan Wang and Chen Bo Calvin Zhang and Noah Jacobson and Bing Liu and Brad Kenstler},
  booktitle = {Proceedings of the 43rd International Conference on Machine Learning},
  year = {2026},
  url = {https://openreview.net/forum?id=uEVTdoAbnK},
}

@inproceedings{sweflow,
  title = {{Synthesizing Software Engineering Data in a Test-Driven Manner}},
  author = {Lei Zhang and Jiaxi Yang and Min Yang and Jian Yang and Mouxiang Chen and Jiajun Zhang and Zeyu Cui and Binyuan Hui and Junyang Lin},
  booktitle = {Proceedings of the 42nd International Conference on Machine Learning},
  volume = {267},
  series = {Proceedings of Machine Learning Research},
  pages = {76518--76540},
  publisher = {PMLR},
  year = {2025},
  url = {https://proceedings.mlr.press/v267/zhang25cn.html},
}

@misc{swehub,
  title = {{SWE-Hub: A Unified Production System for Scalable, Executable Software Engineering Tasks}},
  author = {Yucheng Zeng and Shupeng Li and Daxiang Dong and Ruijie Xu and Zimo Chen and Liwei Zheng and Yuxuan Li and Zhe Zhou and Haotian Zhao and Lun Tian and Heng Xiao and Tianshu Zhu and Longkun Hao and Jianmin Wu},
  year = {2026},
  url = {https://arxiv.org/abs/2603.00575},
  eprint = {2603.00575},
  archivePrefix = {arXiv},
  primaryClass = {cs.AI},
  howpublished = {arXiv preprint arXiv:2603.00575},
}

@misc{swenext,
  title = {{SWE-Next: Scalable Real-World Software Engineering Tasks for Agents}},
  author = {Jiarong Liang and Zhiheng Lyu and Zijie Liu and Xiangchao Chen and Ping Nie and Kai Zou and Wenhu Chen},
  year = {2026},
  url = {https://arxiv.org/abs/2603.20691},
  eprint = {2603.20691},
  archivePrefix = {arXiv},
  primaryClass = {cs.SE},
  howpublished = {arXiv preprint arXiv:2603.20691},
}

@inproceedings{swerebenchv2,
  title = {{SWE-rebench V2: Language-Agnostic SWE Task Collection at Scale}},
  author = {Ibragim Badertdinov and Maksim Nekrashevich and Anton Shevtsov and Alexander Golubev},
  booktitle = {Proceedings of the 43rd International Conference on Machine Learning},
  year = {2026},
  url = {https://openreview.net/forum?id=UCAda9kS57},
}

@inproceedings{swerl,
  title = {{SWE-RL: Advancing LLM Reasoning via Reinforcement Learning on Open Software Evolution}},
  author = {Yuxiang Wei and Olivier Duchenne and Jade Copet and Quentin Carbonneaux and Lingming Zhang and Daniel Fried and Gabriel Synnaeve and Rishabh Singh and Sida I. Wang},
  booktitle = {Advances in Neural Information Processing Systems},
  volume = {38},
  pages = {78500--78525},
  publisher = {Curran Associates, Inc.},
  year = {2025},
  doi = {10.52202/085713-2629},
  url = {https://doi.org/10.52202/085713-2629},
}

@misc{sweshepherd,
  title = {{SWE-Shepherd: Advancing PRMs for Reinforcing Code Agents}},
  author = {Mahir Labib Dihan and Md Ashrafur Rahman Khan},
  year = {2026},
  url = {https://arxiv.org/abs/2604.10493},
  eprint = {2604.10493},
  archivePrefix = {arXiv},
  primaryClass = {cs.SE},
  howpublished = {arXiv preprint arXiv:2604.10493},
}

@inproceedings{swesmith,
  title = {{SWE-smith: Scaling Data for Software Engineering Agents}},
  author = {John Yang and Kilian Lieret and Carlos E. Jimenez and Alexander Wettig and Kabir Khandpur and Yanzhe Zhang and Binyuan Hui and Ofir Press and Ludwig Schmidt and Diyi Yang},
  booktitle = {Advances in Neural Information Processing Systems},
  volume = {38},
  publisher = {Curran Associates, Inc.},
  year = {2025},
  doi = {10.52202/085713-3239},
  url = {https://doi.org/10.52202/085713-3239},
}

@misc{sweuniverse,
  title = {{SWE-Universe: Scale Real-World Verifiable Environments to Millions}},
  author = {Mouxiang Chen and Lei Zhang and Yunlong Feng and Xuwu Wang and Wenting Zhao and Ruisheng Cao and Jiaxi Yang and Jiawei Chen and Mingze Li and Zeyao Ma and Hao Ge and Zongmeng Zhang and Zeyu Cui and Dayiheng Liu and Jingren Zhou and Jianling Sun and Junyang Lin and Binyuan Hui},
  year = {2026},
  url = {https://arxiv.org/abs/2602.02361},
  eprint = {2602.02361},
  archivePrefix = {arXiv},
  primaryClass = {cs.SE},
  howpublished = {arXiv preprint arXiv:2602.02361},
}

@inproceedings{terminalbench,
  title = {{Terminal-Bench: Benchmarking Agents on Hard, Realistic Tasks in Command Line Interfaces}},
  author = {Mike A Merrill and Alexander Glenn Shaw and Nicholas Carlini and Boxuan Li and Harsh Raj and Ivan Bercovich and Lin Shi and Jeong Yeon Shin and Thomas Walshe and E. Kelly Buchanan and Junhong Shen and Guanghao Ye and Haowei Lin and Jason Poulos and Maoyu Wang and Marianna Nezhurina and Di Lu and Orfeas Menis Mastromichalakis and Zhiwei Xu and Zizhao Chen and Yue Liu and Robert Zhang and Leon Liangyu Chen and Anurag Kashyap and Jan-Lucas Uslu and Jeffrey Li and Jianbo Wu and Minghao Yan and Song Bian and Vedang Sharma and Ke Sun and Steven Dillmann and Akshay Anand and Andrew Lanpouthakoun and Bardia Koopah and Changran Hu and Etash Kumar Guha and Gabriel H. S. Dreiman and Jiacheng Zhu and Karl Krauth and Li Zhong and Niklas Muennighoff and Robert Kwesi Amanfu and Shangyin Tan and Shreyas Pimpalgaonkar and Tushar Aggarwal and Xiangning Lin and Xin Lan and Xuandong Zhao and Yiqing Liang and Yuanli Wang and Zilong Wang and Changzhi Zhou and David Heineman and Hange Liu and Harsh Trivedi and John Yang and Junhong Lin and Manish Shetty and Michael Yang and Nabil Omi and Negin Raoof and Shanda Li and Terry Yue Zhuo and Wuwei Lin and Yiwei Dai and Yuxin Wang and Wenhao Chai and Shang Zhou and Dariush Wahdany and Ziyu She and Jiaming Hu and Zhikang Dong and Yuxuan Zhu and Sasha Cui and Ahson Saiyed and Arinbj{\"o}rn Kolbeinsson and Christopher Michael Rytting and Ryan Marten and Yixin Wang and Jenia Jitsev and Alex Dimakis and Andy Konwinski and Ludwig Schmidt},
  booktitle = {{The Fourteenth International Conference on Learning Representations}},
  year = {2026},
  url = {https://openreview.net/forum?id=a7Qa4CcHak},
}

@inproceedings{sweagent,
  title = {{SWE-agent: Agent-Computer Interfaces Enable Automated Software Engineering}},
  author = {Yang, John and Jimenez, Carlos and Wettig, Alexander and Lieret, Kilian and Yao, Shunyu and Narasimhan, Karthik and Press, Ofir},
  booktitle = {Advances in Neural Information Processing Systems},
  volume = {37},
  pages = {50528--50652},
  publisher = {Curran Associates, Inc.},
  year = {2024},
  doi = {10.52202/079017-1601},
  url = {https://doi.org/10.52202/079017-1601},
}

@inproceedings{patchdiff,
  title = {{Are ``Solved Issues'' in SWE-bench Really Solved Correctly? An Empirical Study}},
  author = {You Wang and Michael Pradel and Zhongxin Liu},
  booktitle = {Proceedings of the 2026 IEEE/ACM 48th International Conference on Software Engineering},
  pages = {169--181},
  publisher = {Association for Computing Machinery},
  year = {2026},
  doi = {10.1145/3744916.3764576},
  url = {https://doi.org/10.1145/3744916.3764576},
}

@inproceedings{swebench,
  title = {{SWE-bench: Can Language Models Resolve Real-World GitHub Issues?}},
  author = {Carlos E. Jimenez and John Yang and Alexander Wettig and Shunyu Yao and Kexin Pei and Ofir Press and Karthik Narasimhan},
  booktitle = {The Twelfth International Conference on Learning Representations},
  year = {2024},
  url = {https://openreview.net/forum?id=VTF8yNQM66},
}

@inproceedings{openhands,
  title = {{OpenHands: An Open Platform for AI Software Developers as Generalist Agents}},
  author = {Xingyao Wang and Boxuan Li and Yufan Song and Frank F. Xu and Xiangru Tang and Mingchen Zhuge and Jiayi Pan and Yueqi Song and Bowen Li and Jaskirat Singh and Hoang H. Tran and Fuqiang Li and Ren Ma and Mingzhang Zheng and Bill Qian and Yanjun Shao and Niklas Muennighoff and Yizhe Zhang and Binyuan Hui and Junyang Lin and Robert Brennan and Hao Peng and Heng Ji and Graham Neubig},
  booktitle = {The Thirteenth International Conference on Learning Representations},
  year = {2025},
  url = {https://openreview.net/forum?id=OJd3ayDDoF},
}

@inproceedings{coderl,
  title = {{CodeRL: Mastering Code Generation through Pretrained Models and Deep Reinforcement Learning}},
  author = {Le, Hung and Wang, Yue and Gotmare, Akhilesh Deepak and Savarese, Silvio and Hoi, Steven Chu Hong},
  booktitle = {Advances in Neural Information Processing Systems},
  volume = {35},
  pages = {21314--21328},
  publisher = {Curran Associates, Inc.},
  year = {2022},
  doi = {10.52202/068431-1549},
  url = {https://doi.org/10.52202/068431-1549},
}

@misc{starcoder2,
  title = {{StarCoder 2 and The Stack v2: The Next Generation}},
  author = {Anton Lozhkov and Raymond Li and Loubna Ben Allal and Federico Cassano and Joel Lamy-Poirier and Nouamane Tazi and Ao Tang and Dmytro Pykhtar and Jiawei Liu and Yuxiang Wei and Tianyang Liu and Max Tian and Denis Kocetkov and Arthur Zucker and Younes Belkada and Zijian Wang and Qian Liu and Dmitry Abulkhanov and Indraneil Paul and Zhuang Li and Wen-Ding Li and Megan Risdal and Jia Li and Jian Zhu and Terry Yue Zhuo and Evgenii Zheltonozhskii and Nii Osae Osae Dade and Wenhao Yu and Lucas Krau{\ss} and Naman Jain and Yixuan Su and Xuanli He and Manan Dey and Edoardo Abati and Yekun Chai and Niklas Muennighoff and Xiangru Tang and Muhtasham Oblokulov and Christopher Akiki and Marc Marone and Chenghao Mou and Mayank Mishra and Alex Gu and Binyuan Hui and Tri Dao and Armel Zebaze and Olivier Dehaene and Nicolas Patry and Canwen Xu and Julian McAuley and Han Hu and Torsten Scholak and Sebastien Paquet and Jennifer Robinson and Carolyn Jane Anderson and Nicolas Chapados and Mostofa Patwary and Nima Tajbakhsh and Yacine Jernite and Carlos Mu{\~n}oz Ferrandis and Lingming Zhang and Sean Hughes and Thomas Wolf and Arjun Guha and Leandro von Werra and Harm de Vries},
  year = {2024},
  url = {https://arxiv.org/abs/2402.19173},
  eprint = {2402.19173},
  archivePrefix = {arXiv},
  primaryClass = {cs.SE},
  howpublished = {arXiv preprint arXiv:2402.19173},
}

@article{testoracle,
  title = {{The Oracle Problem in Software Testing: A Survey}},
  author = {Earl T. Barr and Mark Harman and Phil McMinn and Muzammil Shahbaz and Shin Yoo},
  journal = {IEEE Transactions on Software Engineering},
  volume = {41},
  number = {5},
  pages = {507--525},
  year = {2015},
  doi = {10.1109/TSE.2014.2372785},
  url = {https://doi.org/10.1109/TSE.2014.2372785},
}

@inproceedings{evalplus,
  title = {{Is Your Code Generated by ChatGPT Really Correct? Rigorous Evaluation of Large Language Models for Code Generation}},
  author = {Liu, Jiawei and Xia, Chunqiu Steven and Wang, Yuyao and Zhang, Lingming},
  booktitle = {Advances in Neural Information Processing Systems},
  volume = {36},
  pages = {21558--21572},
  publisher = {Curran Associates, Inc.},
  year = {2023},
  doi = {10.52202/075280-0943},
  url = {https://doi.org/10.52202/075280-0943},
}

@inproceedings{codet,
  title = {{CodeT: Code Generation with Generated Tests}},
  author = {Bei Chen and Fengji Zhang and Anh Nguyen and Daoguang Zan and Zeqi Lin and Jian-Guang Lou and Weizhu Chen},
  booktitle = {The Eleventh International Conference on Learning Representations},
  year = {2023},
  url = {https://openreview.net/forum?id=ktrw68Cmu9c},
}

@inproceedings{selfdebugging,
  title = {{Teaching Large Language Models to Self-Debug}},
  author = {Xinyun Chen and Maxwell Lin and Nathanael Sch{\"a}rli and Denny Zhou},
  booktitle = {The Twelfth International Conference on Learning Representations},
  year = {2024},
  url = {https://openreview.net/forum?id=KuPixIqPiq},
}

@inproceedings{reflexion,
  title = {{Reflexion: language agents with verbal reinforcement learning}},
  author = {Shinn, Noah and Cassano, Federico and Gopinath, Ashwin and Narasimhan, Karthik and Yao, Shunyu},
  booktitle = {Advances in Neural Information Processing Systems},
  volume = {36},
  pages = {8634--8652},
  publisher = {Curran Associates, Inc.},
  year = {2023},
  doi = {10.52202/075280-0377},
  url = {https://doi.org/10.52202/075280-0377},
}
\endgroup
\clearpage
\appendix
\renewcommand{\thetable}{\Alph{section}\arabic{table}}
\renewcommand{\theHtable}{appendix.\Alph{section}.\arabic{table}}
\setcounter{table}{0}
\section{Training and Evaluation Configuration}\label{app:training-configuration}

Table~\ref{tab:training-configuration} lists the evaluation setup and shared training settings for the four task pools in Section~\ref{sec:task-scale-quality}.

% Essential training and evaluation settings in one table.
\begin{table}[H]
\centering
\caption{Training and evaluation configuration.}
\label{tab:training-configuration}
\small
\setlength{\tabcolsep}{3pt}
\renewcommand{\arraystretch}{1.0}
\begin{tabularx}{\linewidth}{@{}>{\raggedright\arraybackslash}p{0.55\linewidth}>{\raggedright\arraybackslash}X@{}}
\toprule
\textbf{Setting} & \textbf{Value} \\
\midrule
Initial policy & MiMo-V2.5 \\
Training task pool & 5,545 CodeMidas tasks \\
Algorithm & GRPO \\
Reward & Binary verifier outcome (0 or 1) \\
Advantage normalization by standard deviation & Disabled \\
Batch size & 32 \\
Rollouts per task & 32 \\
Maximum prompt length (tokens) & 8,192 \\
Maximum response length (tokens) & 516,096 \\
Maximum turns per rollout & 500 \\
Maximum staleness & 8 \\
Optimizer & Adam \\
Learning rate & \(5\times10^{-6}\) \\
Warmup steps & 0 \\
Adam \((\beta_1,\beta_2)\) & (0.95, 0.95) \\
Adam \(\epsilon\) & \(10^{-15}\) \\
Gradient clipping threshold & 1 \\
Weight decay & 0 \\
\midrule
CodeMidas Val size & 200 tasks \\
Attempts per task on CodeMidas Val & 3 \\
\bottomrule
\end{tabularx}
\end{table}

\FloatBarrier
\Needspace{14\baselineskip}
\section{Behavioral Metric Definitions}\label{app:analysis}\label{app:trajectory-analysis}

\noindent \textbf{Codebase exploration.} The number of distinct read/search requests before the first codebase edit. Reads with the same target file and line range, and searches with the same query, scope, and options, are counted once. Equivalent requests through dedicated tools or shell commands are treated as duplicates.

\noindent \textbf{Code drafting.} For each \texttt{Write}/\texttt{Edit} payload, we sample distinct 16-character fragments at a stride of four characters. We count fragments found anywhere in reasoning before the corresponding write. The drafting ratio is the summed count divided by the number of fragments sampled across write actions.

\noindent \textbf{Self-verification.} The number of distinct verification commands executed after the final codebase edit, including project test commands, inline checks, temporary test programs, and local program runs. Repeated executions of the same verification command are counted only once.

\FloatBarrier
\end{document}